\documentclass[runningheads]{llncs}
\usepackage[T1]{fontenc}
\usepackage{graphicx}
\usepackage{subcaption}
\usepackage{natbib}
\begin{document}
\title{Real Time Multi Organ Classification on Computed Tomography Images}

\author{Halid Ziya Yerebakan\inst{1} \\ \and
Yoshihisa Shinagawa\inst{1} \\ \and
Gerardo Hermosillo Valadez\inst{1}}

\authorrunning{Yerebakan et al.}
%
\institute{Siemens Medical Solutions 
Malvern, USA\\
\url{http://siemens-healthineers.com} \\
\email{halid.yerebakan@siemens-healthineers.com}}

\maketitle              

\begin{abstract}

Organ segmentation is a fundamental task in medical imaging since it is useful for many clinical automation pipelines. However, some tasks do not require full segmentation. Instead, a classifier can identify the selected organ without segmenting the entire volume. In this study, we demonstrate a classifier based method to obtain organ labels in real time by using a large context size with a sparse data sampling strategy. Although our method operates as an independent classifier at query locations, it can generate full segmentations by querying grid locations at any resolution, offering faster performance than segmentation algorithms. We compared our method with existing segmentation techniques, demonstrating its superior runtime potential for practical applications in medical imaging.

\end{abstract}    

\section{Introduction}
\label{sec:intro}

Medical image segmentation has been a long-standing research topic within the healthcare and medical imaging communities, aimed at enhancing radiology workflows through automation. Compared to landmarking and bounding box detection methods, segmentation provides more granular information. This is beneficial for storing the anatomical locations of findings in structured databases, quantifying abnormalities, radiotherapy planning, calculating doses, comparing longitudinal studies, visualizing imaging data, or filtering the operating region for CAD algorithms. Therefore, it is desirable to have fast and accurate segmentation algorithms in our medical vision pipelines.

The field of medical image segmentation has significantly evolved, transitioning from traditional methodologies to advanced deep learning algorithms and thereby achieving improvements in both the quality and speed of segmentation processes. A notable development in this domain has been the introduction of U-net and its variants, which leverage feature pyramid structures to combine global and local information, enhancing robustness and achieving higher resolution masks \cite{ronneberger2015u}. This approach has not only proved beneficial for medical image segmentation but has also found applicability in a broad range of segmentation tasks across general computer vision.

Recent advancements have introduced transformer architectures to overcome the limitations of convolutional networks' receptive fields, thus enabling superior segmentation outcomes in medical images\cite{hatamizadeh2022unetr,hatamizadeh2021swin}. These methods harness token-to-token interactions across the entire field of view within small patches to capture long-range information. However, the architecture change introduces quadratic complexity in relation to the field of view, posing challenges for computational efficiency.

Independent classifiers are not commonly used for medical image segmentation due to the inefficiency of repetitive sliding window computations and the irregular masks they produce. However, they can quickly obtain organ labels without scanning the entire image. We demonstrate a method to reduce the computational burden of classifier-based segmentation using a data engineering strategy that involves sparse sampling of intensities with a wide field of view in a hierarchical grid, similar to Yerebakan et al. \cite{pointmatching}. This approach allows a classifier to label query locations with less data while maintaining segmentation level accuracy. Applying the classifier to a regular grid at any resolution provides full-volume segmentation, and a refinement operation improves mask granularity at edges. This architecture avoids significant computation and storage costs, making practical applications like storing annotated findings with organ labels feasible. Our experiments on the BTCV dataset show that classifying a single point takes about 0.92 ms, and fine-level segmentation takes around 10 s without additional GPU hardware.

\subsection{Related Work}

Identifying organs in medical images can be approached through various methods. Landmarking techniques provide quick estimates but lack precision for specific organ labeling at arbitrary positions \cite{ghesu2017robust}. Object detection methods define bounding boxes but struggle with precise boundaries. Atlas-based segmentation will be robust thanks to registration but computationally demanding and impractical for real-time applications. Even semantic feature matching strategies fail to achieve real-time processing under 100ms \cite{pointmatching,bai2023sam++,yan2022sam}.

U-Net, introduced by Ronneberger et al. \cite{ronneberger2015u,gibson2018automatic}, has long been a baseline for segmentation algorithms, utilizing multiple image resolutions to balance global and local details. Various studies have modified U-Net to enhance performance \cite{zhou2018unet++}. Our method employs a similar multi-resolution strategy but uses a sampler function to select raw data intensities. This sampling mechanism allows the creation of independent descriptors at voxel locations, allowing rapid point-wise classifications without resampling the image volume.


Recently, transformers have gained popularity in medical imaging. UnetTr \cite{hatamizadeh2022unetr} outperformed U-Net on the BTCV dataset, with further improvements in efficiency and accuracy seen in SwinUnetTR \cite{cao2022swin}. Despite these advancements, computational time remains a challenge for real-time applications, even with GPU hardware.


Foundational models, similar to Large Language Models in NLP, are gaining traction in medical vision. Vox2Vec \cite{goncharov2023vox2vec} offers unsupervised learning from image data, while other studies focus on pretraining self-supervised models for better segmentation accuracy \cite{chen2023masked}. These methods create voxel-level vectors for classification, but still require processing large subvolumes even for a single point.


The datasets and pretrained models for medical image segmentation are expanding. One notable dataset for classification is MedMNIST \cite{yang2023medmnist}, which serves as an educational benchmark. However, its real-world application is limited due to the varying size of regions of interest. Another notable dataset, the Total Segmentator Dataset, includes 1,204 images of 117 organs \cite{wasserthal2023totalsegmentator}, and the model is available in the Monai library. The algorithm runs in about 30 seconds on a GPU and several minutes on a CPU. Competitions like the Flare challenge aim for fast segmentation, with top methods reported to achieve around 10 seconds runtime on a GPU \cite{myronenko2023automated}. This demonstrates the gap in utilizing segmentation methods for user-interfacing medical imaging applications.


\nocite{chen2023masked}
\nocite{tadokoro2023pre}
\nocite{azad2024beyond}

\section{Our Method}

Our method consists of two steps. First, we perform multi-resolution sparse sampling around the point of interest and then apply a deep neural network classifier to the collected descriptor. This process assigns an organ label to the selected point in the volumetric image.

\subsection{Sparse Sampling Image Intensities}
\begin{figure}[ht]
    \centering
    \begin{subfigure}[b]{0.45\linewidth}
        \centering
        \includegraphics[width=\linewidth]{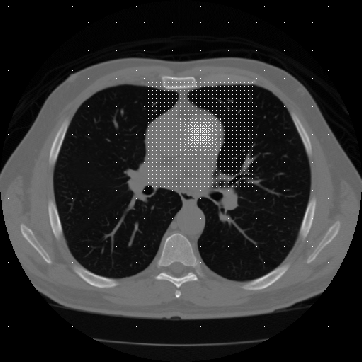}
        \caption{Sampling Offsets For Descriptor}
        \label{fig:sampling}
    \end{subfigure}
    \hfill
    \begin{subfigure}[b]{0.45\linewidth}
        \centering
        \includegraphics[width=\linewidth]{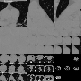}
        \caption{Decoding Image From A Descriptor}
        \label{fig:decoding}
    \end{subfigure}
    \caption{Descriptor Definition and Decoding}
    \label{fig:descriptor_definition}
\end{figure}

The resampling step is very common in medical image segmentation pipelines. Despite being basic, it is still time-consuming for interactive use and requires extra memory. Instead of resampling the image, our method defines a sampler function with fixed millimeter offsets. Then, descriptors are simply the image intensities at those offset locations. This sampler works only on selected query points, significantly reducing the required computation for any point of interest in the image.

The organ label at a point of interest depends on the intensity values near the point itself. Additionally, context is crucial for disambiguating similar-looking soft tissue types. Therefore, instead of sampling with a single resolution and a limited subvolume, we use different regular grids at various resolutions to capture a large area hierarchically. Thanks to the consistency of the human body, similar anatomical locations generate similar descriptors. The fixed offsets function like distance sensors, encoding location.

The sampler offsets are in millimeters and are adjusted to voxel offsets once the image is loaded, by dividing by voxel spacing and rounding. If the offset locations are outside the image volume, they are assigned a value of 0 for the corresponding dimension, ensuring all descriptors are in the same vector space. During runtime, descriptor computation involves only memory lookup operations. Memory positions are determined by adding offsets to the current voxel.

In our experiments, we utilized 2D+3D grids for sampling. The 2D part consists of 3 orthogonal planes defined at a 4mm resolution by a 27x27 grid. The 3D part comprises six three-dimensional grids at resolutions of 2, 3, 5, 12, 28, and 64 mm, respectively, with 9x9x9 grid sizes, considering some slice thicknesses in CT images. The spacing increases with non-integer multipliers to avoid sampling the same locations in different resolution grids. The selection of a size 9 grid allows both 2D and 3D to be present in the same picture, simplifying debugging. We named this approach 3.5D sampling since it contains more than 3D information.

Exemplary sampling grids in a single slice given a point are shown in Figure \ref{fig:sampling}, where bright dots are the sampling locations. The hierarchical nature of the sampling allows for a very large field of view with granular detail in the center. These descriptors can be decoded back into an 81x81 2D image as seen in Figure \ref{fig:decoding}, by placing every 27x27 block in 9 different places. In this figure, the upper left corner is reserved for the 2D portion of the descriptor with three orthogonal planes, and the other parts are 3D slices at different resolutions with 9x9x9 grids. The total dimension of the descriptor becomes 6561. As seen in the figure, the query location belongs to the heart.

While other descriptor definitions are possible, for the sake of simple decoding for visualization, we used this 9x9x9x9 descriptor definition described here. The organ label is defined as the segmentation label at the query point.

\subsection{Classification}

\begin{figure}
\centering
\includegraphics[width=0.7\linewidth]{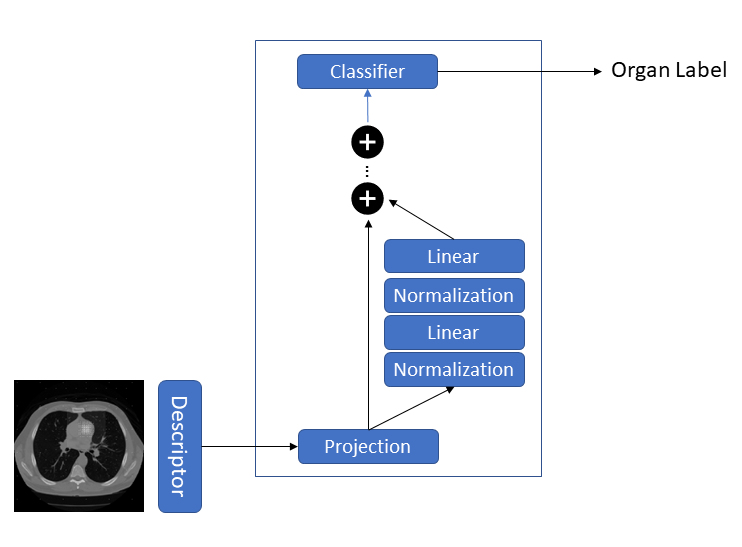}
\caption{1D Residual Network on Sampled Intensities}
\label{fig:classifier}
\hfill
\end{figure}

We have chosen a 1D residual feedforward network as the classifier in this work. This approach is faster because there is only one forward pass per sampled point, and it does not have translation invariance since translation will change the organ labels in the center of sampling. The classifier outputs the organ label probabilities given the sparsely sampled flat 1D descriptor at that point. Residual connections are implemented using two-layer linear blocks interleaved with normalization. After the initial projection, we use the same number of hidden dimensions until the final classifier layer. Since the full descriptor is passed as a 1D input, all resolutions are fused in the middle layers. The model architecture is visualized in Figure \ref{fig:classifier}. Swish activation functions are used after the linear layers. Projection is the most computational part of the classifier due to 6561$\times$nhidden number of parameters where nhidden is smaller than the initial descriptor size.





\subsection{Segmentation}

The classifier provides functionality to obtain the organ label for any query location. Thus, it is possible to compute a segmentation mask by independently querying all voxels. However, querying every location in the image is computationally expensive. Instead, we can use a hierarchical grid with multiple resolutions to obtain faster results.

\begin{figure}[ht]
\centering
\begin{subfigure}[b]{0.32\linewidth}
    \centering
    \includegraphics[width=\linewidth]{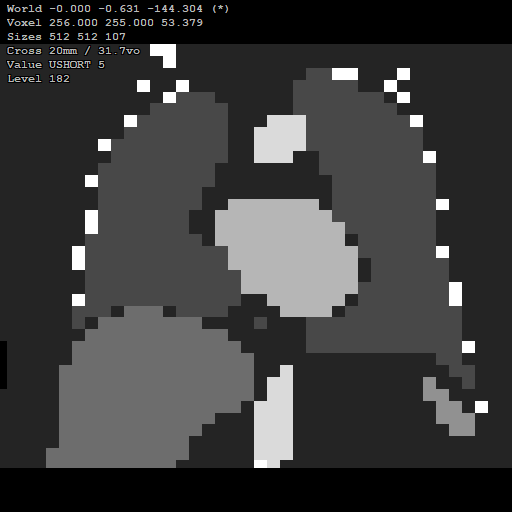}
    \caption{Coarse Segmentation}
    \label{fig:coarse}
\end{subfigure}
\hfill
\begin{subfigure}[b]{0.32\linewidth}
    \centering
    \includegraphics[width=\linewidth]{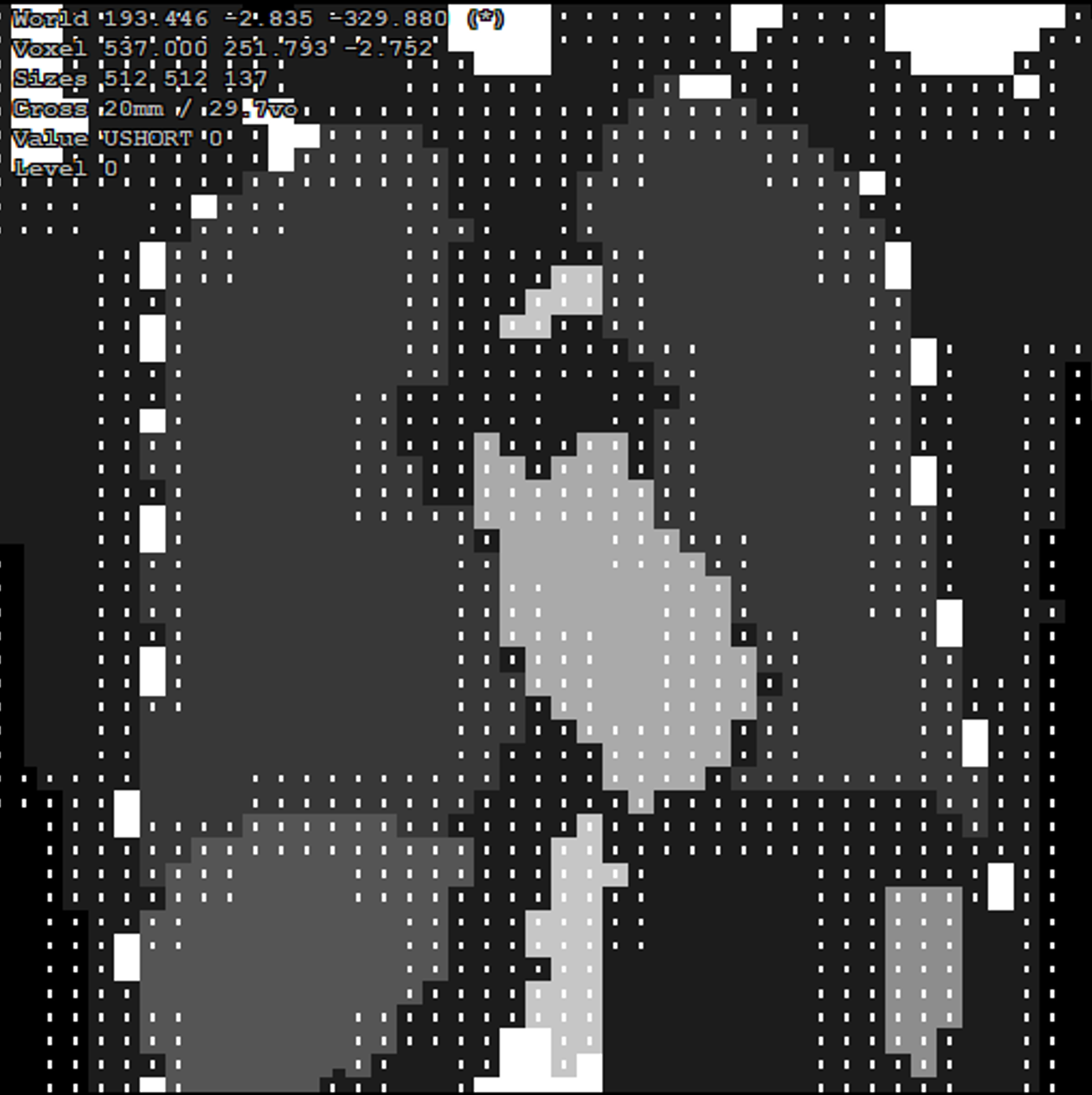}
    \caption{Edge Refinement}
    \label{fig:refinement}
\end{subfigure}
\hfill
\begin{subfigure}[b]{0.32\linewidth}
    \centering
    \includegraphics[width=\linewidth]{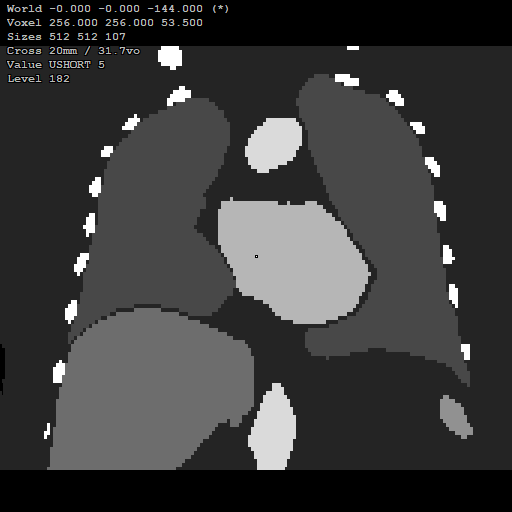}
    \caption{Fine Details}
    \label{fig:fine}
\end{subfigure}
\caption{Classifier based segmentation enables segmentation in any resolution thus allowing multiple steps of refinement for fine segmentation}
\end{figure}

We used a high-level 8mm sparse grid to create a coarse segmentation mask by filling the 8mm subvolumes with predicted labels as shown in an example in Figure \ref{fig:coarse} where each intensity represents a different organ label. However, masks are not accurate near edges due to low resolution. Since errors are happening on the edges, we could further refine the resolution in those regions hierarchically. To achieve this, we have checked all points in the segmentation mask with higher resolution grid and added points into the task queue to query if the neighbors have different labels, as shown in Figure \ref{fig:refinement}. For each level, we have some smoothing on the edges by thresholding a majority of 20 within 27 neighbors to prevent excessive computation. Thus, the cost of classification in high resolution at nearly homogeneous points would be eliminated. We reduced the spacing to 2mm with this approach. 

An example generated mask is shown in Figure \ref{fig:fine}. Thanks to points being completely independent of each other, there is a massive parallelization potential.


\section{Experiments}

\begin{figure*}[ht]
	\centering
	\begin{subfigure}[b]{0.4\textwidth}
		\centering
		\includegraphics[width=\linewidth]{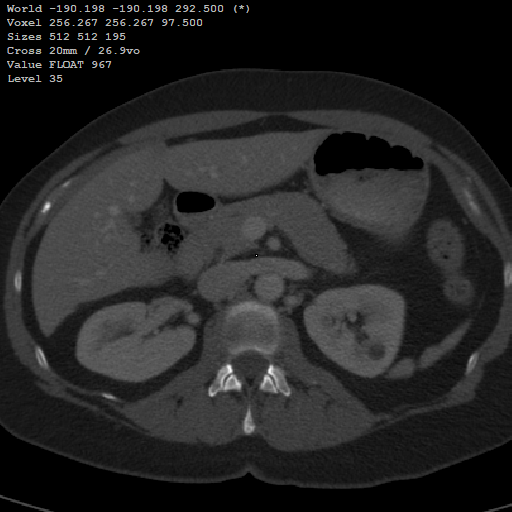}
		\caption{Image Slice}
		\label{fig:imageslice}
	\end{subfigure}
	\hfill 
	\begin{subfigure}[b]{0.4\textwidth}
		\centering
		\includegraphics[width=\linewidth]{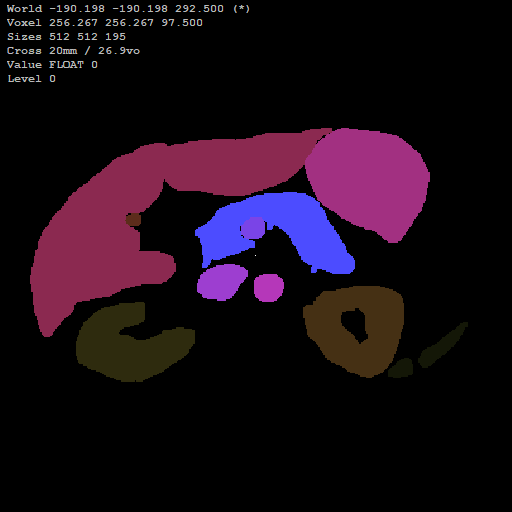}
		\caption{Ground Truth Mask}
		\label{fig:coarseresult}
	\end{subfigure}
	
	\vspace{1cm} 
	
	\begin{subfigure}[b]{0.4\textwidth}
		\centering
		\includegraphics[width=\linewidth]{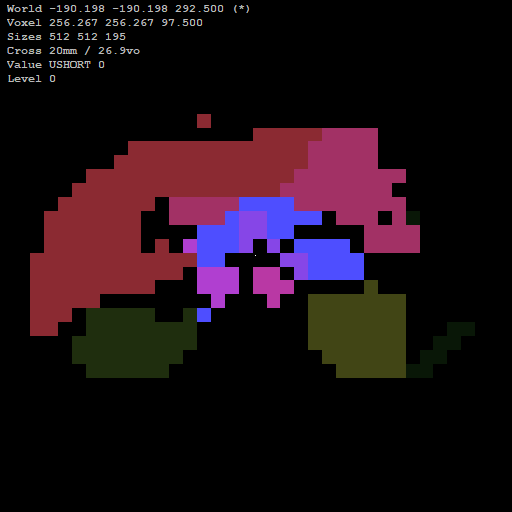}
		\caption{Coarse Segmentation}
		\label{fig:refineresult}
	\end{subfigure}
	\hfill 
	\begin{subfigure}[b]{0.4\textwidth}
		\centering
		\includegraphics[width=\linewidth]{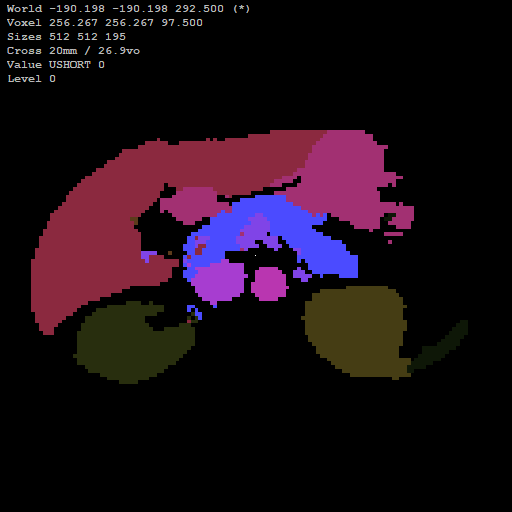}
		\caption{Fine Mask}
		\label{fig:gt}
	\end{subfigure}
	
	\caption{Qualitative Results on BTCV dataset}
	\label{fig:results}
\end{figure*}

We have experimented with our method on the public BTCV dataset, which contains a total of 30 images with 13 abdominal organ labels. We trained with 24 cases and used 6 cases for validation, following the official split for evaluation, similar to other studies. The data is accessible online \footnote{https://www.synapse.org/\#!Synapse:syn3193805/wiki/217789}.

For the training dataset generation, we have sampled random locations globally and random locations from each class. The latter provides some balancing to labels, which could be adjusted. In our experiments, we have used 100k examples per training image, of which 10\% belong to the balanced set. The corresponding classifier label is obtained from the center voxel of the mask image. The image descriptors are divided by 128 and clipped between -4 and 4. 

We have used 8 layers of neural network with 128 dimensions mostly considering performance constraints. Our initial learning rate is 3e-4, with cosine decay reducing the learning rate down to 1e-6. L1-L2 regularization is used with 1e-5 coefficient in addition to 0.25 drop out ratio. The model is trained for 100 epochs on standard cross-entropy loss. We did not add any data augmentation. In fact, random flips are against the positional priors of the human body if images are loaded correctly, especially in the abdomen region, but small random rotations could be helpful. 

Once the model is trained, it is converted to C++ counterpart to reduce latency. We have used OpenMP parallelization for different query points for segmentation part of the experiment. Time has measured on 6 core - 12 thread Xeon CPU. 

In the classification part, we had 1000 calls to validation set in random locations and compared against ground truth. Total time is measured for each case, and average time is calculated. The average time per query is 0.92±0.01 milliseconds. With this speed, we are able to obtain 97.4\% accuracy with a macro average 86.76\% F1 score, including background class in the classification task. The classifier speed is similar across different volumes since volume size does not have a direct impact on descriptor size.

In the segmentation part, we first obtained coarse segmentation for full volume. An example 2D slice is shown in Figure \ref{fig:coarseresult}. The average coarse segmentation speed is 5.24±0.9s per volume. Segmentation is also refined in the edges using the same classifier. The speed, in this case, is an average of 9.51±2.72s on the CPU. We measured dice scores and compared them against recently published studies on BTCV dataset. In terms of runtime speed, Shaker et. al. \cite{shaker2022unetr++} reports GPU time of 62.4 seconds while being the most efficient algorithm therein. The results are shown in Table \ref{tab:multi_organ_segmentation}.

\begin{table*}[ht]
	\centering
	\caption{Dice Score Averages on Multi Organ Segmentation on BTCV validation}
	\label{tab:multi_organ_segmentation}
	\resizebox{\textwidth}{!}{%
		\begin{tabular}{l|ccccccccccccc|c}
			\hline
			Methods & Spleen & RKid & LKid & Gall & Eso & Liv & Sto & Aor & IVC & Veins & Pan & RAG & LAG & AVG \\
			\hline
			\textbf{From Scratch} & \\
			ViT3D-B \cite{chen2023masked}      & 0.8902 & 0.8926 & 0.8769 & 0.4763 & 0.4891 & 0.9447 & 0.7475 & 0.8207 & 0.773  & 0.6175 & 0.6442 & 0.5663 & 0.4699 & 0.7084 \\
			nnFormer \cite{shaker2022unetr++}  & 0.9458 & 0.8862 & 0.9368 & 0.6529 & 0.7622 & 0.9617 & 0.8359 & 0.8909 & 0.8080 & 0.7597 & 0.7787 & 0.7020 & 0.6605 & 0.8162 \\
			UnetTR++ \cite{shaker2022unetr++}  & 0.9494 & 0.9190 & 0.9362 & 0.7075 & 0.7718 & 0.9595 & 0.8515 & 0.8928 & 0.8314 & 0.7691 & 0.7742 & 0.7256 & 0.6817 & 0.8328 \\
			\hline
			\textbf{Pretrained} & \\
			ViT3D-L\cite{chen2023masked} & 0.9556 & 0.9582 & 0.9414 & 0.5206 & 0.5352 & 0.9898 & 0.8025 & 0.8811 & 0.8298 & 0.6649 & 0.6916 & 0.6088 & 0.5045 & 0.7603 \\
						
			\hline
			\textbf{Ours} & 0.8398 & 0.8364 & 0.7990 & 0.4780 & 0.6489 & 0.9264 & 0.6860 & 0.7770 & 0.7106 & 0.5007 & 0.5583 & 0.4054 & 0.4089 & 0.6596 \\
			\hline
		\end{tabular}
	}
\end{table*}

Our algorithm, on average, reaches a dice score of 65.96 without any augmentation or additional training data. Despite being lower as compared to full volume segmentation this value defines a new benchmark for a new category of point classification based segmentation methods.

\section{Discussion}
It is possible to improve the proposed system in various ways. We list some of them in this section. 

A connected component analysis could be further included to remove false positive errors in the segmentation masks before refinement. The biggest connected component could be selected to remove erroneous regions before further refining the edges. This method is likely to improve both the speed and the accuracy of the system. 

The independence of the classifier is a limitation in segmentation operation. It is possible to create an additional filter on extracted class probabilities further to reduce the variance within local estimations. Another alternative is utilizing independent classifier features as token embeddings for further transformer layers instead of processing full 3D raw data. However, in this case, real-time response needs to be sacrificed. The power of an independent classifier is that it works on any resolution and any region of interest without the need for full-volume processing. 

\section{Conclusion}

We have demonstrated an efficient multi-organ classification algorithm that can produce organ labels in real time. This algorithm could be utilized further to create segmentation masks for full 3D CT volumes within seconds without additional hardware. Despite the loss in accuracy, the runtime advantage makes it amenable to many practical use cases. The proposed data selection strategy's advantages extend beyond enhancing segmentation efficiency; it holds the potential to accelerate a variety of medical imaging tasks, including object detection, registration, and landmarking. In the future, we are considering experimenting with more organs and alternative deep learning architectures. 
\bibliographystyle{plain}
\bibliography{FastSegmentation}
\end{document}